\documentclass{article}

\usepackage{PRIMEarxiv}

\usepackage[utf8]{inputenc} % allow utf-8 input
\usepackage[T1]{fontenc}    % use 8-bit T1 fonts
\usepackage{hyperref}       % hyperlinks
\usepackage{url}            % simple URL typesetting
\usepackage{booktabs}       % professional-quality tables
\usepackage{amsfonts}       % blackboard math symbols
\usepackage{nicefrac}       % compact symbols for 1/2, etc.
\usepackage{microtype}      % microtypography
\usepackage{lipsum}
\usepackage{fancyhdr}       % header
\usepackage{graphicx}       % graphics
\usepackage{amsmath}
\usepackage{amsfonts}
\usepackage{amssymb}
\usepackage{algorithm}
\usepackage{algpseudocode}
\usepackage{subcaption}
\usepackage{subcaption}
\usepackage{amsmath,bm}

\graphicspath{{media/}}     % organize your images and other figures under media/ folder

\title{New Probabilistic-Dynamic Multi-Method Ensembles for Optimization based on the CRO-SL
}

\author{
Jorge P\'erez-Aracil \\
Signal Theory and Communications\\ Universidad de Alcal\'a, Spain \\
\texttt{jorge.perezaracil@uah.es} \\
 \And
Carlos Camacho-G\'omez \\
Computer Systems Department, \\
Universidad Polit\'{e}cnica de Madrid, Spain\\
\texttt{carlos.camacho@upm.es} \\
\\
\And
Eugenio Lorente-Ramos\\
Signal Theory and Communications\\ Universidad de Alcal\'a, Spain \\
\texttt{eugenio.lorente@edu.uah.es} \\
\And
Cosmin M. Marina\\
Signal Theory and Communications\\ Universidad de Alcal\'a, Spain \\
\texttt{cosmin.marina@edu.uah.es} \\
\\
\And
Sancho Salcedo-Sanz\\
Signal Theory and Communications\\ Universidad de Alcal\'a, Spain \\
\texttt{sancho.salcedo@uah.es} \\
}

\begin{document}
\maketitle

\begin{abstract}
In this paper we propose new probabilistic and dynamic (adaptive) strategies to create multi-method ensembles based on the Coral Reefs Optimization with Substrate Layers (CRO-SL) algorithm. The CRO-SL is an evolutionary-based ensemble approach, able to combine different search procedures within a single population. In this work we discuss two different probabilistic strategies to improve the algorithm. First, we defined the Probabilistic CRO-SL (PCRO-SL), which substitutes the substrates in the CRO-SL population by {\em tags} associated with each individual. Each tag represents a different operator which will modify the individual in the reproduction phase. In each generation of the algorithm, the tags are randomly assigned to the individuals with a similar probability, obtaining this way an ensemble with a more intense change in the application of different operators to a given individual than the original CRO-SL. The second strategy discussed in this paper is the Dynamical Probabilistic CRO-SL (DPCRO-SL), in which the  probability of tag assignment is modified during the evolution of the algorithm, depending on the quality of the solutions generated in each substrate. Thus, the best substrates in the search process will be assigned with a higher probability that those which showed a worse performance during the search. We test the performance of the proposed probabilistic and dynamic ensembles in different optimization problems, including benchmark functions and a real application of wind turbines layout optimization, comparing the results obtained with that of existing algorithms in the literature.
\end{abstract}

% keywords can be removed

\section{Introduction}
\label{sec:sample1}

In optimization problems, an {\em ensemble} method refers to an algorithm that combines different types of alternative methods, search strategies or operators, in order to obtain high-quality solutions \cite{wu2019ensemble}. The application of ensemble approaches has been massive in the last few years, due to the good results obtained by these combinations of techniques in hard optimization problems and real applications. Following \cite{wu2019ensemble}, there are different types of ensemble approaches: high-level ensembles, focused on selecting the best optimization algorithm for a given problem, and low-level ensembles, which refers to an optimal combination of different types of search strategies or operators within a single approach. In any case, the main idea of ensemble algorithms is to exploit the capacity of different methods by combining them in several possible ways, in order to improve the search ability of the final approach in optimization problems.

It is possible to find different ensemble algorithms recently proposed in the literature, including multi-method and multi-strategy approaches. Multi-method algorithms consider the combination of different operators or algorithms to solve an optimization problem. An example of low-level competitive single population approaches is \cite{vrugt2007improved}, in which different operators are applied in a single evolutionary algorithm-based ensemble. An approach with a similar idea was proposed in \cite{vrugt2008self}. Multi-method approaches have also been applied to improve the performance of meta-heuristics in multi-objective optimization problems \cite{mashwani2016multiobjective}. There are also multi-method algorithms which work on different sub-populations such as \cite{xue2014ensemble}. Following this idea, in \cite{price2022animorphic} an ``animorphic ensemble optimization'' is presented, where a set of algorithms form an ensemble, demonstrating stronger performance over different problems than each components on their own. This approach exploits the concept of islands in algorithms, in such a way that several populations, each based on different search approaches are defined. An island model interface strategy is then defined, where populations exchange of individuals is promoted, depending on the performance of the algorithm associated with each population. 
There are also high-level ensembles which combine operators with different strategies, such as \cite{peng2010population}, where a portfolio of different algorithms for optimization problems is proposed. Note that high-level ensemble are close to the general idea of hyper-heuristics \cite{drake2020recent}, in this case the objective is to choose the best combination of algorithms depending on the problem tackled. In \cite{grobler2013multi} a high-level multi-strategy ensemble choose among different meta-heuristics such as evolutionary algorithms, Particle Swarm Optimization (PSO) algorithm or Evolutionary Strategies. Ensembles of multi-strategy approaches in which different versions of the same operator are chosen can be also found in the literature, for example, PSO \cite{du2008multi}, Artificial Bee Colony algorithms \cite{wang2014multi} or Biogeography-based optimization \cite{xiong2013multi}. Note that alternative versions of optimization ensembles may involve other algorithmic components (not only different search operators), such as neighborhood sizes \cite{zhao2012decomposition} or constraint handling
techniques \cite{mallipeddi2010ensemble}, among others. 

In consonant with the previous discussion, note that one of the most successful meta-heuristic to construct optimization ensembles and multi-strategy algorithms using variants of the same technique is Differential Evolution \cite{das2010differential}. Ensembles and multi-strategy algorithms based on Differential Evolution (DE) started to appear over one decade ago. One of the first ensemble and multi-strategies approaches which merged different variants of DE were \cite{gong2011adaptive} and \cite{mallipeddi2011differential}. In \cite{gong2011adaptive} two DE variants with adaptive strategy selection were proposed. The idea was that the algorithm autonomously select the most suitable strategy while solving the problem, according to their recent impact on the optimization process. In turn, \cite{mallipeddi2011differential} proposed an ensemble of DE-based mutation strategies, together with control parameters, which are forced to coexists in a single population, and throughout the evolution process they compete to produce the best possible offspring. In \cite{awad2018ensemble} a DE ensemble based on the specific algorithm's version called LSHADE \cite{tanabe2014improving}, with an ensemble parameter sinusoidal adaptation (LSHADE-EpSin) was proposed. In \cite{ali2016adaptive} a  multi-population ensemble based on {\em tribes of DE versions} was introduced. In that approach, the population is clustered in multiple tribes, and uses an ensemble of different mutation and crossover strategies. A competitive success-based scheme is applied to determine the contribution of each tribe to the next generation of the ensemble. The approach was successfully tested in CEC2014 benchmark suits. In \cite{wu2016differential} an ensemble of multiple DE strategies based on a multi-population scheme was proposed. Specifically, three DE mutation strategies where tested: ``current-to-pbest/1'', ``current-to-rand/1'' and ``rand/1'' as part of the DE ensemble. Results in CEC 2005 benchmark functions were reported to be competitive to other meta-heuristic approaches for continuous optimization problems. In \cite{wu2018ensemble} a multi-population based DE ensemble, called Ensemble of Differential Evolution Variants (EDEV), was proposed to obtain a efficient algorithm for real encoding optimization problems. Recently, in \cite{yao2021improved} the EDEV approach was revisited and improved. In \cite{li2019two} a two-stage ensemble of DE variants for numerical optimization was proposed. This ensemble approach was based on two different stages. In the first one, a multi-population approach is used, which includes three different DE variants (SHADE, JADE, and DE/current-to-rand/1). In a second stage of the algorithm,  LSHADE is used to improve the convergence of the algorithm. This approach were tested in functions from the CEC2005 benchmark suits. In \cite{wang2021shade} another ensemble involving two DE versions were proposed. Specifically, two versions of the L-SHADE approach, L-SHADE-EpSin and L-SHADE-RSP, were selected and inter-changed during the searching process forming an ensemble approach with two basic methods, in order to improve the results in real-encoded optimization problems. 

Recently, a multi-method ensemble known as Coral Reefs Optimization with Substrate Layer (CRO-SL) was proposed \cite{Sancho_IJBIC,salcedo2016coral,salcedo2017review}, and successfully applied to very different optimization problems in Science and Engineering fields such as: energy and microgrids design \cite{salcedo2016novel,jimenez2019optimal,perez2022versatile}, mechanical and structural design \cite{salcedo2017structures,camacho2018active,perez2020submerged,hernandez2022hybridizing,perez2021eliminating}, electrical engineering \cite{sanchez2018optimal,camacho2019coral,camacho2020design}. The CRO-SL is a low-level, evolutionary-based multi-method ensemble, which combines different types of search operators within a single population (reef), by dividing it in different zones (substrates), in which a different operator is applied. The evolution of the population is then carried out by applying the different operators to the population, depending on the zone in which a solution is located. A given solution may be formed by combination of other solutions in the population with a given operator at a time (2 points crossover, multi-point crossover, differential evolution, etc.) or modified with mutation-based operators (Gaussian mutation, chaotic-based, Cauchy mutations, etc.), which may also formed part of the methods implemented in the ensemble. The new solutions are settled in the population at random locations, which promotes the application of different operators in the evolution. The number and type of methods included in the CRO-SL is a decision of the practitioner, and must be defined previously to the ensemble run.

Albeit the CRO-SL has obtained notable success when tackling different optimization problems, in this paper we revisit the algorithm, and propose new adaptive strategies to improve the algorithm's design, in order to make more efficient the way in which the different operators are applied in the ensemble. Thus, in this work we first discuss two different adaptive (probabilistic) strategies for the CRO-SL dynamics modification. The first one, called Probabilistic CRO-SL (PCRO-SL), substitutes the zones (substrates) in the CRO-SL population by {\em tags} associated with each individual. Each tag represents then a different operator which will modify the individual in the reproduction phase. In each generation of the ensemble, the tags are randomly assigned to the individuals with a similar probability, obtaining this way an ensemble with a more intense change in the application of different operators to a given individual. The second strategy proposed to improve the CRO-SL is called Dynamical Probabilistic CRO-SL (DPCRO-SL), and in this case we keep the tag assignment of evolution methods to each individual, but the probability of assignment is modified during the evolution of the algorithm, depending on the quality of the solutions generated in each substrate. Thus, those substrates which obtain better results up to a given point in the search process, will be assigned with more probability that those which performed worse during the search. Note that this process tries to promote the evolution with operators which obtain good results, and reducing the evolution with other operators which are not contributing to the generation of good solutions to the problem. We evaluate the different proposed versions of the CRO-SL-based multi-method ensemble (PCRO-SL and DPCRO-SL), in a large set of benchmarks instances, and in a real optimization problem of wind turbine layouts, considering different sets of substrates. We will compare the probabilistic and dynamical versions of the CRO-SL against the classical CRO-SL version, and also with alternative meta-heuristics previously published in the literature.

The rest of the paper has been structured in the following way: next section presents the original CRO-SL ensemble and the theoretical basis of different substrates used, such as different versions of Differential Evolution, Firefly algorithm, Two-points crossover, BLX-$\alpha$ crossover or Gaussian and Cauchy-based mutations. Section \ref{sec:PCRO-SL} presents the new Probabilistic Adaptive CRO-SL proposed in this work. Section \ref{sec:Experiments} presents the experiments and results obtained with the new multi-method ensemble proposed. Finally, Section \ref{sec:Conclusions} closes the paper with some conclusions and remarks on the research carried out.

\section{Methods}\label{sec:methods}

In this section the basic approaches which have been used are described. First, the CRO-SL is shown. Then the most important characteristics of the commonly used heuristics and meta-heuristics included as substrates in the CRO-SL algorithm are described.

\subsection{The CRO-SL: a multi-method ensemble evolutionary algorithm}\label{CRO-SL}

The Coral Reefs Optimisation algorithm with Substrate Layers (CRO-SL) \cite{Sancho_IJBIC,salcedo2016coral} is a low-level ensemble for optimisation \cite{wu2019ensemble}, based on evolutionary computation. It was first proposed as an advanced version of a basic original algorithm CRO \cite{salcedo2014coral}. We describe the CRO-SL multi-method ensemble here, starting by introducing the basic CRO approach first. 

\subsubsection{Basic CRO}\label{basic_CRO}

The Coral Reef Optimization Algorithm (CRO) \cite{salcedo2014coral,salcedo2017review} is an evolutionary-type meta-heuristic, proposed as a class of hybrid between Evolutionary Algorithms \cite{del2019bio} and Simulated Annealing \cite{kirkpatrick1983optimization}. The original CRO uses a model of a rectangular-shaped reef of size $M \times N$, ($\Lambda$), where the possible solutions to the problem at hand (corals) are set. Each space $\Lambda (i,j)$, where $i$ and $j$ are the space's coordinates, can be empty or contain a coral ${\bf x}_k$. The algorithm carries out an evolution of the solutions in the reef, as follows:

\begin{enumerate}
    \item \textbf{Initialization}: A fraction $\rho_0$ of the total reef capacity is occupied with randomly generated corals. The reef position that each coral occupies is also randomly selected.
    \item \textbf{Evolution}: Once the reef has been populated the evolution process begins. This process is divided into five phases per generation:
    \begin{enumerate}
        \item \textbf{Sexual reproduction}: In this phase, new solutions (larvae set) are created from the ones belonging to the reef in order to compete for a place in the reef. Sexual reproduction can be performed in two ways: external and internal. A percentage $F_b$ of the corals settled in the reef performs external reproduction (Broadcast spawning) and the rest of them $(1-F_b)$ reproduce themselves through internal sexual reproduction (Brooding). Both reproduction processes are performed as follows:
            \begin{enumerate}
                \item \textbf{Broadcast spawning}: from the set of corals selected for external sexual reproduction ($F_b$), new solutions (larvae) are generated and released.
                \item \textbf{Brooding}: each one of the remaining corals ($1-F_b$) produces a larva by means of a small perturbation and releases it.
            \end{enumerate}
        \item \textbf{Larvae setting}: In this step, all the larvae produced by Broadcast spawning or Brooding try to find a spot in the reef to grow up. A reef position is randomly chosen, and the larva will settle in that spot only in one of the following scenarios:
        \begin{enumerate}
            \item The spot is empty.
            \item The larva has a better health function value (fitness) than the coral currently occupying that spot.
        \end{enumerate}
        Each larva can try to settle in the reef a maximum of three times. If the larva has not been able to settle down in the reef after that number of attempts, it is discarded.
        \item \textbf{Asexual reproduction}: In this phase (also called budding) a fraction $F_a$ of the corals with better fitness present in the reef duplicate themselves and, after a small mutation, are released. They will try to settle in the reef as in the previously described step.
        \item \textbf{Depredation}: Finally, each coral belonging to the $F_{dep}$ worst fraction can be predated (erased from the reef) with a low probability $P_d$.
    \end{enumerate}
\end{enumerate}

This basic version of the algorithm works as an evolutionary-type approach, defined in exploitation, not in exploration, as the majority of algorithms do. This means that we can use any kind of search procedure in the CRO. In fact, the first algorithm in \cite{salcedo2014coral} used a 2-points crossover operator to perform the Broadcast spawning, but in other cases alternative operators were considered, such as Harmony Search operators \cite{salcedo2015coral} or $\beta$-hillclimbing \cite{ahmed2021improved}. Note that this paves the way to define an improved algorithm as a multi-method ensemble.

\subsubsection{CRO with Substrate Layers (CRO-SL)}

\begin{figure}[!ht]
    \centering
    \includegraphics[draft=false, angle=0,width=10cm]{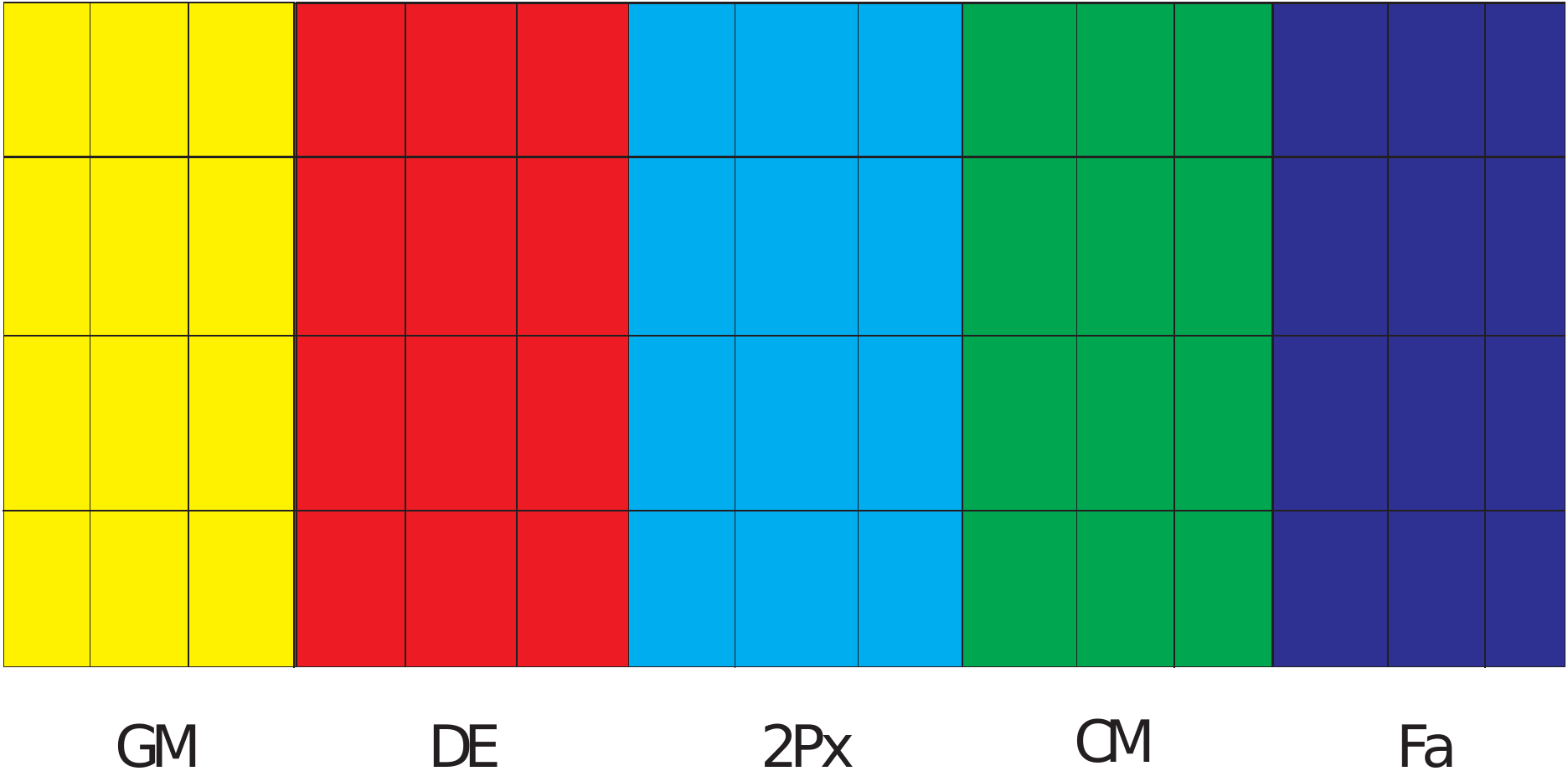}
    \caption{Reef in the CRO-SL example. An example where 5 different substrates stand for different search procedures applied: Gaussian Mutation (yellow); Differential Evolution (red); Two-points crossover (light blue); Cauchy mutation (green) and Firefly algorithm (dark blue).}
    \label{crosl_reef}
\end{figure}

The CRO-SL algorithm \cite{Sancho_IJBIC,salcedo2016coral} is a further evolution of the CRO approach towards a multi-method ensemble. It generally proceeds as the basic CRO, but with a significant difference: instead of having a single surface of size $M \times N$, it considers several {\em substrate layers} ($T$) of the approximately same size in the reef (Figure \ref{crosl_reef}). Each substrate, in turn, represents a particular evolution strategy or searching procedure. Thus, the CRO-SL is a multi-method ensemble algorithm \cite{wu2019ensemble}, where several searching strategies are carried out within a single population.

Figure \ref{crosl_reef} shows a visual description of the CRO-SL procedure. This new approach adds a dimension to the reef $\Lambda$, so a reef position is now given by three coordinates $\Lambda(t,i,j)$, where $t$ is the substrate index, and $i$ and $j$ have the same meaning as in basic CRO. In Figure \ref{crosl_reef} the third dimension is represented with colours. Thus, the evolutionary process is the same as in the basic CRO at a general level, but the reproduction phase is performed at substrate level, so that a different search operator is applied depending on the substrate the solution is allocated. The Brooding phase remains the same as the basic CRO, for all substrates. The produced larvae are released to a common reservoir, and then the larvae setting procedure is carried out as in the basic CRO, regardless of its original substrate. 

\subsection{Substrate layers defined in the CRO-SL}\label{sec:substrates}
Very different search strategies can be defined in the CRO-SL as part of the multi-method approach, and they affect the performance of the ensemble. They are usually defined at the practitioner's discretion. In related articles, different combinations of well-known meta-heuristics have been defined. In this case we again test regular combinations of previously-defined heuristics and meta-heuristics, depending on the problem at hand. Specifically, we have defined and applied the following substrates in the CRO-SL and its new variants:
Differential Evolution (DE) (different versions), Firefly algorithm (Fa), classical two-points crossover (2Px), BLX-$\alpha$ crossver (BLX), Gaussian-based mutation (GM) and Cauchy-based mutation (CM).

\begin{enumerate}

\item DE: Differential Evolution (DE) algorithm \cite{storn1997differential} is a stochastic population-based method, specifically designed for global optimization problems \cite{leon2014investigation}. In its more common form, DE maintains a population with $N_p$ individuals, where every individual within the population stands for a possible solution to the problem. Individuals are represented by a vector $X_{i,g}$, where $i=1,\ldots,N_p$ and $g$ refers to the index of the generation. A normal DE cycle consists of three consecutive steps: mutation, crossover and selection. We adapt the algorithm for the CRO-SL by considering only the mutation and crossover parts of the meta-heuristic. Thus, mutation is carried out to generate random perturbations on the population. For each individual, a mutant vector is generated. There are different approaches for DE mutation in the literature \cite{storn1997differential}. We describe here the procedure known as ``best mutation strategy'' \cite{xu2012differential}, which has been successfully applied in many optimization problems before. It attempts to mutate the best individual of the population, according to Equation \eqref{eq_DE1}, where $V_{i,g}$ denotes the mutated vector, $i$ is the index of the vector, $g$ stands for the generation index, $r_1,r_2\in {1,\ldots,N_p}$ are randomly created integers, $X_{best,g}$ denotes the best solution in the population and $F$ is the scaling factor in the interval $[0,2]$. This mutation strategy uses the scaled difference between two randomly selected vectors to mutate the best individual in the population. 

\begin{equation}
\label{eq_DE1}
    V_{i,g}=X_{best,g}+F \cdot (X_{r1,g}-X_{r2,g})
\end{equation}

A crossover procedure is then applied between the mutated vector created in the mutation stage and a randomly chosen individual in the population. The new solutions created are called trial vectors and denoted by $T_{i,g}$ for individual $i$ at generation $g$. Every parameter in the trial vector are decided following Equation \eqref{eq_DE2}, where $j$ represents the index of every parameter in a vector, $CR$ is the probability of recombination, and $J_{rand}$ denotes a randomly selected integer within $(1,\ldots,N_p)$ to ensure that at least one parameter from mutated vector enters the trial vector:
 
\begin{equation}
\label{eq_DE2}
T_{i,g}[j]=\left\{ \begin{array}{ll}
    V_{i,g}[j] & \text{if}\quad rand[0,1]<CR\quad \text{or} \quad j=j_{rand}\\
    X_{i,g}[j] & \text{otherwise}
\end{array} \right.
\end{equation}

% Figure \ref{fig_DE1} shows how a new mutant vector is obtained with this strategy, where $d$ denotes the difference vector between $X_{r1,g}$ and $X_{r2,g}$. 
% \begin{figure}[!ht]
%     \centering
%     \includegraphics[clip,width=0.5\textwidth]{DE_1.pdf}
%     \caption{Best mutation $i$ DE with one difference vector.}
%     \label{fig_DE1}
% \end{figure}

\item Firefly Optimization (Fa): The Fa is a kind of swarm intelligence algorithm, based on the flashing patterns and behaviour of fireflies in nature \cite{yang2009firefly,yang2020firefly}. In this algorithm, the pattern movement of a firefly $i$ attracted to another (brighter) firefly $j$ is calculated as follows:

\begin{equation}\label{firefly_movement}
{\bf x}_i^{t+1}={\bf x}_i^t+\beta_0 e^{-\gamma r_{ij}^2} ({\bf x}_j^t - {\bf x}_i^t)+ \alpha \boldsymbol{\epsilon}_i^t
\end{equation}
where $\beta_0$ stands for the attractiveness at distance $r=0$.  The specific Fa mutation implemented in the CRO-SL is a modified version of the algorithm known as Neighbourhood Attraction Firefly Algorithm (NaFa) \cite{wang2017firefly}. It has been implemented as follows: when a coral (solution) in the reef belongs to the Fa substrate, it is updated following Equation (\ref{firefly_movement}). All the parameters of the equation are tuned during the CRO-SL evolution. The corals in the Fa substrate consider as swarm a neighbourhood among all other corals in the reef (not only the Fa substrate). Thus, the corals in the Fa substrate are updated taking into account some solutions from other substrates, since all the corals in the reef share the same objective function. 

% \item HS: Mutation from the Harmony Search algorithm. Harmony Search~\cite{geem2001new,manjarres2013survey} is a population based meta-heuristic that mimics the improvisation of a music orchestra while its composing a melody. HS controls how new larvae are generated in one of the following ways: i) with a probability HMCR$\in[0,1]$ (Harmony Memory Considering Rate), the value of a component of the new larva is drawn uniformly from the same values of the component in the other corals. ii) with a probability PAR$\in[0,1]$ (Pitch Adjusting Rate), subtle adjustments are applied to the values of the current larva, replaced by any of its neighbouring values (upper or lower, with equal probability).
% %

\item 2Px: Classical 2-points crossover. The crossover operator is the most classical exploration mechanism in genetic and evolutionary algorithms \cite{eiben2003introduction,del2019bio}. It consists of coupling individuals at random, choosing two points for the crossover, and interchanging the genetic material in-between both points. In the classical version of the CRO-SL, one individual to be crossed is from the 2Px substrate, whereas the couple can be chosen from any part of the reef.
%
% \item MPx: Multi-points crossover. Similar to the 2-points crossover, but in this case a number $k$ of crossover points are selected, and a binary template decides whether parts of the individuals are interchanged.
% %

\item BLX: BLX-$\alpha$ crossover. This crossover operator \cite{herrera2002multiple} considers two real-encoded vectors  $x_1 = (x_{11},\ldots, x_{n1})$ and $x_2 = (x_{12},\ldots, x_{n2})$ and generates two offspring, $h^k = (h_{1}^k,\ldots, \delta_{i}^k,\ldots,  h_{n}^k)$, $k=1,2$, where $\delta_{i}^k$ is a randomly (uniformly) chosen number from the interval $\left[x_{min}-I\alpha, x_{max}+I\alpha \right]$, where  $x_{max} = \max{(x_{i1}, x_{i2})}$, $x_{min} =\min{(x_{i1}, x_{i2})}$, and $I = x_{max}-x_{min}$.

\item GM: Gaussian Mutation, with a $\sigma$ value linearly decreasing during the run, from $0.2 \cdot (A-B)$ to $0.02 \cdot (A-B)$, where $[B,A]$ is the domain search. Specifically, the Gaussian probability density function is:
\[
f_{G(0,\sigma^2)}(x)={\frac {1}{\sigma {\sqrt {2\pi }}}}e^{-{\frac {x^2}{2\sigma^2}}}.
\]
The reason of adapting the value of $\sigma$ along the generations is to provide a stronger mutation in the beginning of the optimization, while fine tuning with smaller displacements nearing the end. The mutated larva is thus calculated as: $x'_i=x_i+\delta N_i(0,1)$, where $N_i(0,1)$ is a random number following the Gaussian distribution.

\item CM: Cauchy Mutation. The one-dimensional Cauchy density function centered at the origin is defined by:

\begin{equation}
    f_t(x)=\frac{1}{\pi}\frac{t}{t^2+x^2}
\end{equation}
where $t>0$ is a scale parameter \cite{yao1999evolutionary}, in this case $t=1$. Note that the Cauchy probability distribution looks like the Gaussian distribution, but it approaches the axis so slowly that an expectation does not exist. As a result, the variance of the Cauchy
distribution is infinite \cite{yao1999evolutionary}. In this case, the mutated larva is calculated as: $x'_i=x_i+\eta \delta$, where $\eta$ stands for a variance, and $\delta$ is a random number following the Cauchy distribution.

\end{enumerate}

\section{Proposed Probabilistic-Dynamic ensembles with the CRO-SL}\label{sec:PCRO-SL}

In this section we present the two newly proposed multi-method ensembles from the CRO-SL. Note that the main contribution of these new ensembles is the way that each search procedure is selected for offspring generation, in such a way that it only affects to the broadcast spawning process previously defined. In the original CRO-SL algorithm, each search procedure is assigned to a set of positions of the population (substrate). Thus, every individual settled on one of these positions will follow the same search method in every iteration. Now, in these new CRO-SL versions, the search procedures are chosen dynamically for each parent in each iteration of the run. This means that the search procedures are no longer tied to a set of positions, but a coral   will produce the offspring at each iteration following one of the search procedures, randomly chosen. The main difference between both versions is whether the probabilities are fixed and maintained during the algorithm's run, or they changed dynamically according to the search procedure's performance.

\subsection{Probabilistic CRO-SL ensemble}

\begin{figure}[!ht]
    \centering
    \includegraphics[draft=false, angle=0,width=8cm]{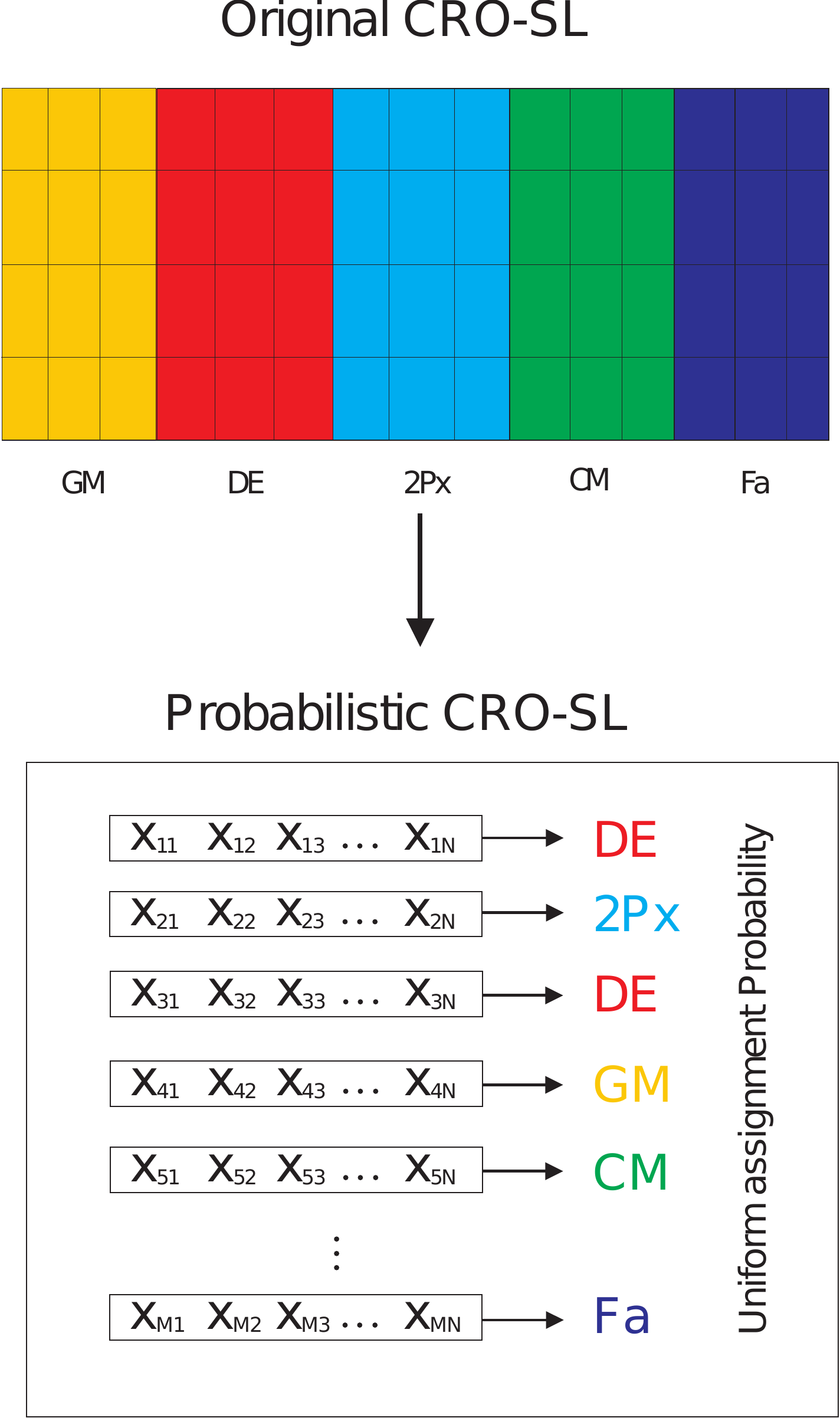}
    \caption{Proposed PCRO-SL, and comparison with the original CRO-SL version.}
    \label{PCRO-SL}
\end{figure}

The Probabilistic CRO-SL ensemble (PCRO-SL) is constructed from the base of the original CRO-SL, by changing the substrates structure for a tag associated with each coral (solution) in the reef. Each tag $t$ stands for the substrate index in this case. The main difference with the CRO-SL is that in each generation, the assignment of tags to corals is changed, so for a given coral, the search procedure changes in each generation according to a given probability distribution, usually uniform. Figure \ref{PCRO-SL} shows an example of the PCRO-SL, comparing the reef with the original CRO-SL.

In essence, the PCRO-SL lets the search methods be independent of the positions in the reef, which still has $M \times N$ size. Now, the substrates are not defined by specific positions in the reef, but they would be formed by a set of individuals randomly distributed throughout the reef. The probability of using one or another search methods by an individual in any iteration of the run is defined by Equation \eqref{eq:PCROprob},

\begin{equation}
    \label{eq:PCROprob}
    \centering
    p_i=1/T
\end{equation}
where $p_i$ stands for the probability that an individual belongs to substrate $i$, and $T$ stands for the number of search procedures (substrates) considered. Note that, in this case, the probabilities $p_i$ do not change during the run, however, the assignment of individuals to each substrate is carried out every generation. Algorithm \ref{alg:PCRO} shows the pseudo-code of this PCRO-SL version.

\begin{algorithm}
    \caption{Probabilistic CRO-SL} 
    \label{alg:PCRO}
    \textbf{Input}: values of the algorithm parameters within the range, including the probabilities of each search method.
    \textbf{Output}: the fittest solution found for the problem at hand.
	\begin{algorithmic}
    	\item Step 1: set the initial population and empty positions, and calculate their fitness values.
    	\item Step 2: each individual can create new solutions in two ways:
    	\If{$F_b$}
    	    \State with a high probability $F_b$ it is generated offspring by the broadcast spawning process: in this version, one search method is randomly selected (with probability $p_i$, given by Equation \eqref{eq:PCROprob}) among the candidates. 
    	\Else
    	    \State with a low probability $(1- F_b)$ it is generated offspring by the brooding process.
    	\EndIf
    	\State Step 4: perform the settlement of the offspring.
    	\State Step 5: with probability $P_d$ the depredation process is carried out.
    	\State Step 6: return the optimal solution if the stopping criterion is hold or go back to step 2 otherwise.
	\end{algorithmic} 
\end{algorithm}

\subsection{Dynamic Probabilistic CRO-SL ensemble}\label{DPCRO-SL_Ensemble}

The PCRO-SL ensemble  described above can be improved by including a dynamic procedure of method probability assignment, in such a way that the most efficient methods have a larger probability to be assigned than other search approaches which have not been so good during the search. Note that there are different possibilities to carry out this dynamical assignment. Specifically, we have evaluated three different ways of calculating the probability of the search method to be assigned to corals in the reef:

\begin{itemize}
\item[1.]{Larvae success rate metric.}
The first probability assignment procedure depends on the rate of success of the larvae (new solutions) produced by the corals in each substrate. In words, during the larvae setting phase, we keep track of the substrate (search method) from which each larva was produced, and we annotate the amount of them that were successful in being inserting into the reef. The new probability of searching methods is obtained as the rate of these successes to the total amount of generated larvae.\\

\item[2.]{Raw fitness metric.}
The second probability assignment procedure is set by the fitness 
 of the generated solutions, i.e. we consider the quality of individual solutions to obtain a metric to each substrate. In words, if the operator applied generates ``good'' solutions, it will have a higher probability of being assigned as search method in future assignments. Note that there are different ways of implementing this metric: for example, we can take the average of the fitness of all the larvae produced, the best fitness across all of them or the worst one, etc.

\item[3.]{Improvement of fitness.}
The last procedure for assigning the methods probabilities is a differential approach, based on the difference with the best fitness obtained in the previous generation. It works
very similarly to the previous strategy, giving a higher value to those substrates that generated solutions with a better fitness.
This method also allows some variants, so we can take the average of the difference, the best or worst values to assign the final probability of method assignment.
\end{itemize}

\begin{figure}[!ht]
    \centering
    \includegraphics[draft=false, angle=0,width=8cm]{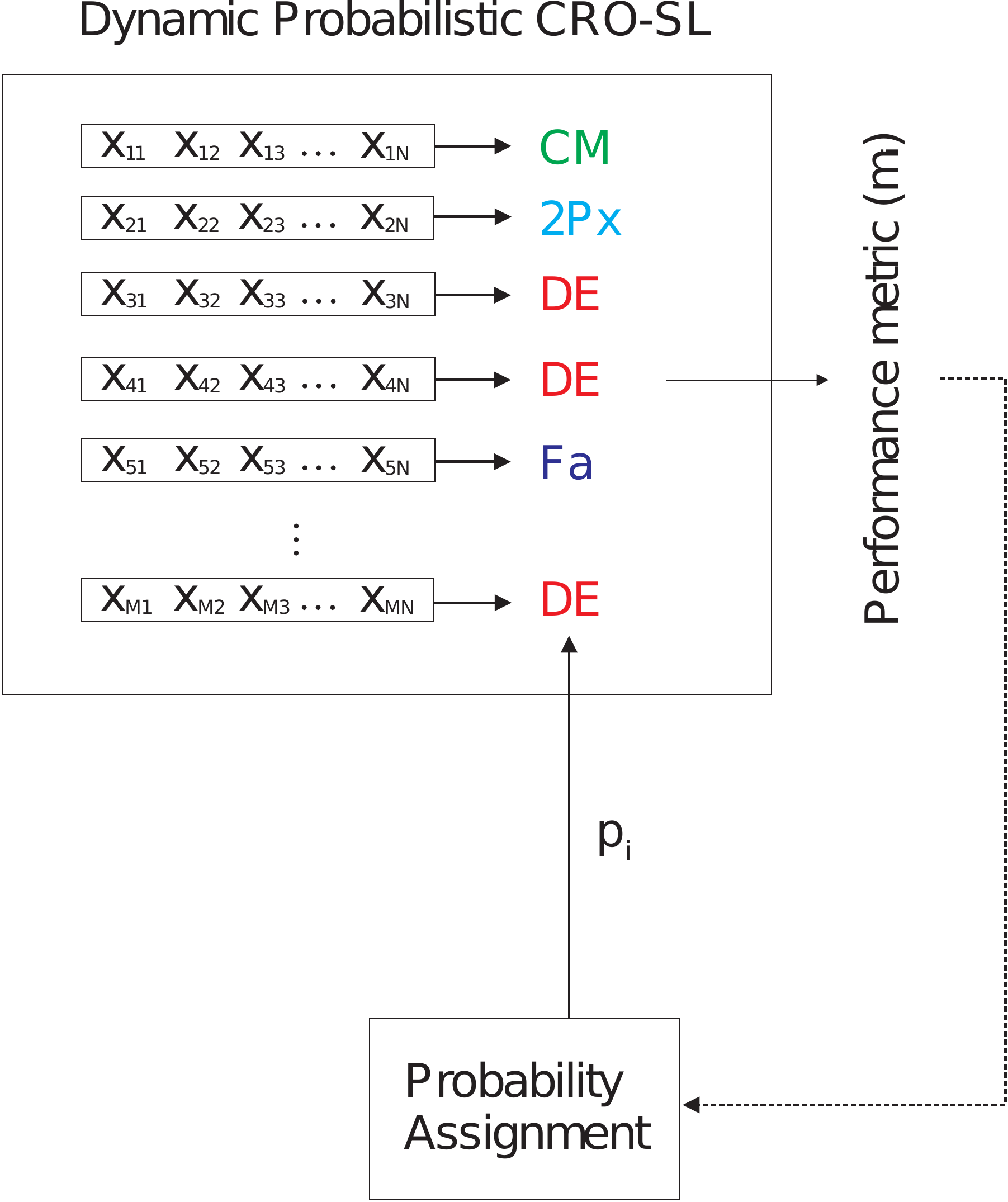}
    \caption{Outline of the dynamic probabilistic CRO-SL algorithm (DPCRO-SL).}
    \label{DPCRO-SL}
\end{figure}

Once we have evaluated each substrate, we can generate a probability distribution from the metric considered, to finally assign the probability for a given substrate in the next generation. To do this, we use the Softmax function, so the probability assigned to one of the $T$ substrates $i$ with a metric $m_i$ can be calculated as follows:

\begin{equation}
\label{eq:DPCROprob}
p_i = \frac{e^{m_i/\tau}}{\sum_{j=0}^{S} e^{m_j/\tau}}
\end{equation}
where the parameter $\tau$ gives a way of ``amplifying'' the probabilities, i.e. making similar changes in the metric of each substrate give higher probabilities with a lower value of $\tau$.
Note that this process of new probability assignment is carried out after a number of generations, $\mathcal{T}$, large enough so we can evaluate the performance of the different search methods in the problem at hand. Figure \ref{DPCRO-SL} shows an outline of the DPCRO-SL ensemble.

To ensure that we are exploring effectively the space of operators, we set a probability threshold 
$\varepsilon$ so that all substrates have at least a probability $\varepsilon$ of being assigned
to an individual, this probability can be very small, but enough to ensure that the operator will be 
eventually chosen. Algorithm \ref{alg:DPCRO} shows the pseudo-code of the DPCRO-SL version.

\begin{algorithm}
    \caption{Dynamic Probabilistic CRO-SL} 
    \label{alg:DPCRO}
    \textbf{Input}: values of the algorithm parameters within the range.
    \textbf{Output}: the fittest solution found for the problem at hand.
	\begin{algorithmic}
    	\item Step 1: set the initial population and empty positions, and calculate their fitness values.
    	\item Step 2: each individual can create new solutions in two ways:
    	\If{$F_b$}
    	    \State with a high probability $F_b$ it is generated offspring by the broadcast spawning process: in this version, one search method is randomly selected (with probability $p_i$ given by Equation \eqref{eq:DPCROprob}) among the candidates.
    	\Else
    	    \State with a low probability $(1 - F_b)$ it is generated offspring by the brooding process.
    	\EndIf
    	\State Step 4: perform the settlement of the offspring.
    	\State Step 5: calculate the success ratio of each search method and update the probabilities following Equation \eqref{eq:DPCROprob}.
    	\State Step 6: with probability $P_d$ the predation process is carried out.
    	\State Step 7: return the optimal solution if the stopping criterion is hold or go back to step 2 otherwise.
	\end{algorithmic} 
\end{algorithm}

%%%%%%%%%%%%%%%%%%%%%%%%%%%%%%%%%%%%%%%%%%
\section{Experimental results}\label{sec:Experiments}

The evaluation of the proposed CRO-SL variants will be carried out in different benchmark functions and also in a real application of wind turbines layout. 

\subsection{Comparison in Benchmark functions}
In this section we compare the performance of the proposed PCRO-SL and DPCRO-SL with the original CRO-SL and other state-of-the-art algorithms in different benchmark functions, to evaluate the goodness of the two newly proposed probabilistic CRO-SL ensembles. The definition of the 15 benchmark functions considered can be found in the Appendix section. In a first set of experiments, we run the PCRO-SL and DPCRO-SL considering as search methods the combination of 4 DE approaches in the ensemble. The reason for defining a DE-based ensemble for the experiments with benchmark functions is that DE-based approaches have obtained excellent results in the past in these kind of problems, such as the LSHADE approach \cite{wang2021shade}. For each of the 15 benchmark functions 10 times with a limit of $3\cdot 10^{5}$ evaluations of the objective functions is considered, and we get the best average and standard deviation across the 10 executions carried out.

The defined DE-based CRO-SL is a version of the algorithm in which we restrict the operators to be used in each substrate to a variant of the cross operation in the DE (differential evolution) algorithm (see Section \ref{sec:substrates}). To define a DE variant, the notation is usally ''DE/a/b'', where \textbf{a} determines which vectors we are going to choose, and \textbf{b} determines how many differences we are going to be calculated. Hence, the variant DE/rand/2 will take 5 vectors at random from the population $X_{r1, g}, X_{r2, g}, X_{r3, g}, X_{r4, g}, X_{r5, g}$ and will calculate the vector $V$ as:
$$V_{i, g} = X_{r1, g} + F \cdot (X_{r2, g} - X_{r3, g}) + F \cdot (X_{r4, g} - X_{r5, g})$$
which will be crossed with the individual chosen in the same way as in the base DE algorithm.\\

\noindent
In these experiments over benchmark functions, we will first use the following DE variants:
\begin{itemize}
    \item[1.] DE/best/1
    $$V_{i, g} = X_{best, g} + F \cdot (X_{r1, g} - X_{r2, g})$$
    \item[2.] DE/best/2
    $$V_{i, g} = X_{best, g} + F \cdot (X_{r1, g} - X_{r2, g}) + F \cdot (X_{r3, g} - X_{r4, g})$$
    \item[3.] DE/current-to-best/1
    $$V_{i, g} = X_{i, g} + U \cdot (X_{best, g} - X_{i, g}) + F \cdot (X_{r1, g} - X_{r2, g})$$
    \item[4.] DE/current-to-pbest/1
    $$V_{i, g} = X_{i, g} + F \cdot (X_{pbest, g} - X_{i, g}) + F \cdot (X_{r1, g} - X_{r2, g})$$
\end{itemize}
Where $U$ is a random value following an uniform probability distribution between 0 and 1, $X_{best,g}$ 
is the individual with the best fitness in generation $g$, $X_{pbest,g}$ is a solution picked at random from the $p\%$ best ones in the generation, $X_{i, g}$ is the individual chosen to be 
crossed with and $X_{rn, g}$ is an individual chosen at random from the population.

\begin{figure}[!ht]
    \centering
    \begin{subfigure}{0.35\linewidth}
        \includegraphics[width=\linewidth]{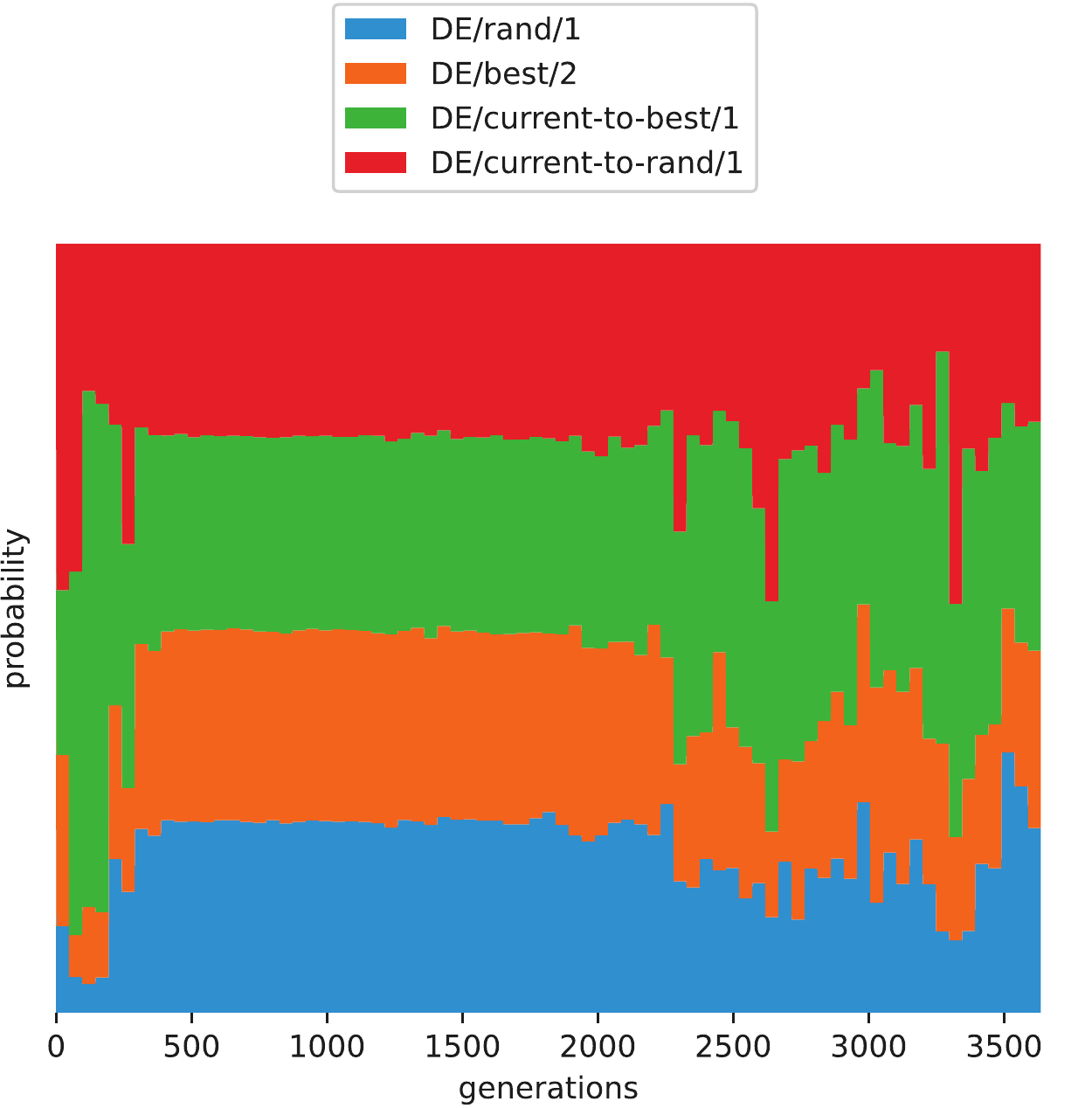}
        \caption{Larvae raw fitness probability assignment method.}
    \end{subfigure}
    \vspace{15pt}
    \begin{subfigure}{0.35\linewidth}
        \includegraphics[width=\linewidth]{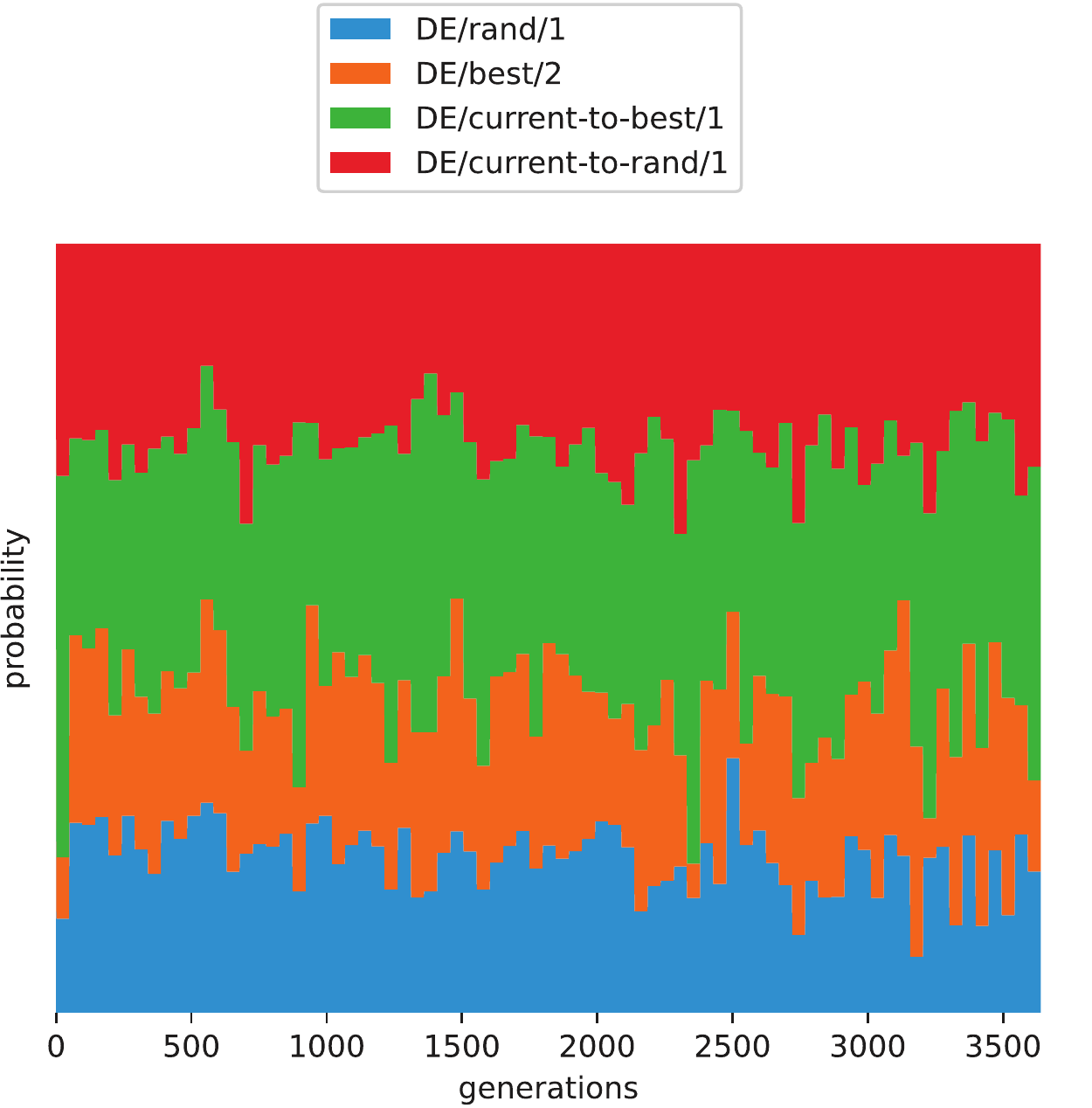}
        \caption{Larvae fitness improvement probability assignment method.}
    \end{subfigure}
    \begin{subfigure}{0.35\linewidth}
        \includegraphics[width=\linewidth]{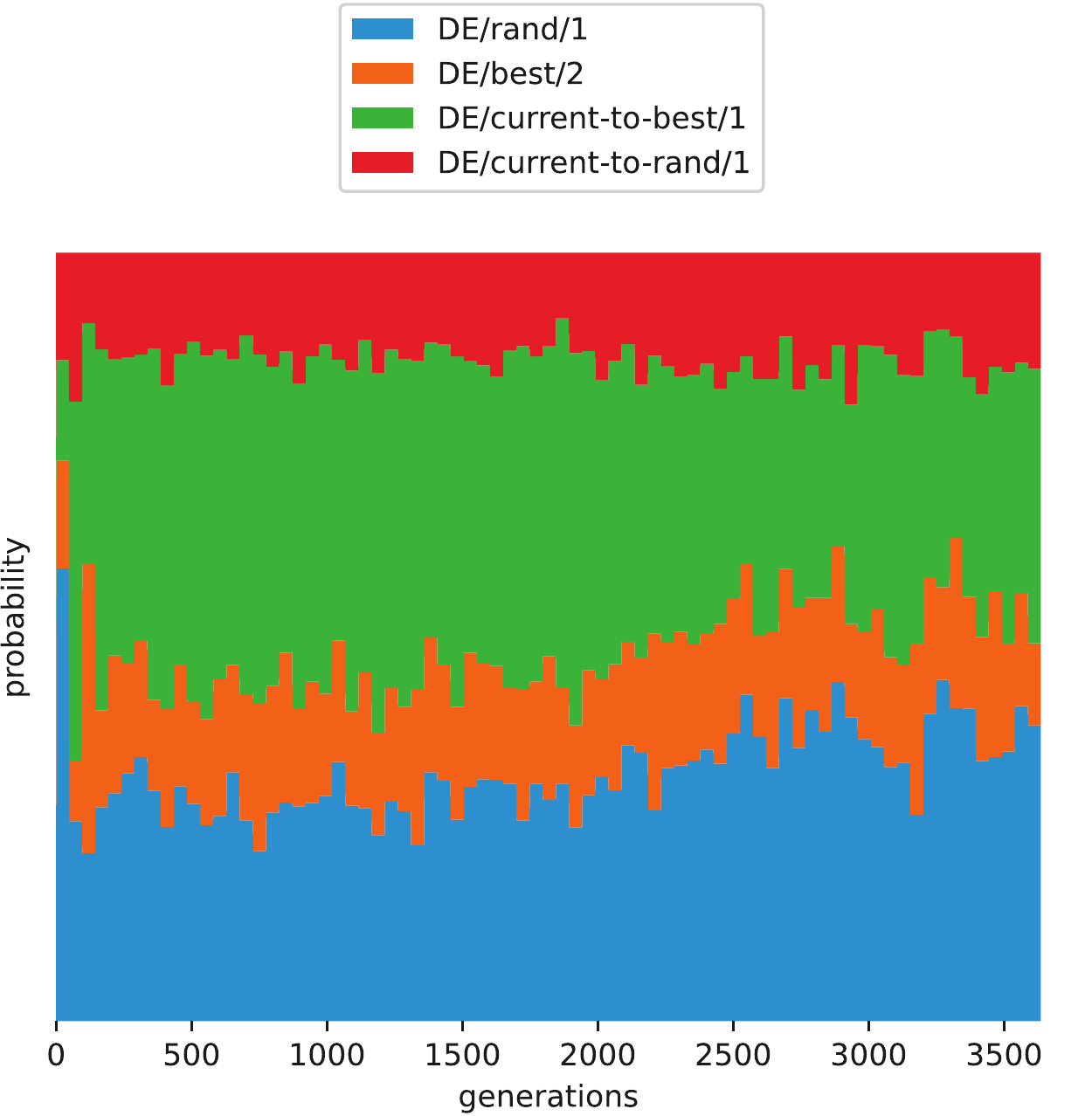}
        \caption{Larvae success rate probability assignment method.}
    \end{subfigure}
     \caption{Probability assignment methods performance in F5 (Rosenbrock benchmark function). The probability is depicted in a relative plot so the colour thickness represents the probability of a given search method assignment. Note that the sum of the assignment probabilities in each generation must be 1.}
     \label{fig:DPCRO_assign_prob}
\end{figure}

Before further testing the performance of the proposed ensembles, we first proceed to evaluate the different methods of probability assignment proposed. Figures \ref{fig:DPCRO_assign_prob} (a), (b) and (c) compare different methods of probability assignment in the DPCRO-SL (raw fitness assignment, fitness improvement and larvae success rate, respectively, described in Section \ref{DPCRO-SL_Ensemble}). Note that the probability is depicted in these figures in a relative plot fashion, so the colour thickness represents the probability of a given search method assignment.  This is an example for the optimization of the Rosenbrock function (F5) with a limit of $3 \cdot 10^5$ evaluations of the function. The probability that is assigned to each operator each generation of the algorithm is shown in the figures. It is possible to see differences in the probability assignment process. After some experimental tests, the best results were obtained with the raw fitness probability assignment process. The rest of the results in these benchmark functions were therefore obtained with this probability assignment method.

% \begin{figure}[!h]
%     \centering
%     \includegraphics[width=0.8\textwidth]{dyn_plots/prob_fbest2.pdf}
%     \caption{DPCRO-SL with the raw fitness method}
%     \label{fig:DPCROevalfbest}
% \end{figure}

% \begin{figure}[!h]
%     \centering
%     \includegraphics[width=0.8\textwidth]{dyn_plots/prob_dbest2.pdf}
%     \caption{DPCRO-SL with the fitness improvement method}
%     \label{fig:DPCROevaldbest}
% \end{figure}

% \begin{figure}[!h]
%     \centering
%     \includegraphics[width=0.8\textwidth]{dyn_plots/prob_succ2.pdf}
%     \caption{DPCRO-SL with the larvae success rate method}
%     \label{fig:DPCROevalsucc}
% \end{figure}

% \begin{figure}[!h]
%     \centering
%     \begin{subfigure}{0.4\linewidth}
%         \includegraphics[width=\linewidth]{dyn_plots/fit_dbest.pdf}
%         \caption{Larvae raw fitness probability assignment method.}
%     \end{subfigure}
%     \vspace{15pt}
%     \begin{subfigure}{0.4\linewidth}
%         \includegraphics[width=\linewidth]{dyn_plots/fit_fbest.pdf}
%         \caption{Larvae fitness improvement probability assignment method.}
%     \end{subfigure}
%     \begin{subfigure}{0.4\linewidth}
%         \includegraphics[width=\linewidth]{dyn_plots/fit_succ.pdf}
%         \caption{Larvae success rate probability assignment method.}
%     \end{subfigure}
%      \caption{Probability assignment methods performance in F5 (Rosembrock benchmark function).}
%      \label{fig:DPCRO_fit_evol}
% \end{figure}

Table \ref{tab:tableCROVAR} shows the results obtained in the optimization of bechmark functions with the different CRO-SL approaches proposed, and the original one. As can be seen, the new PCRO-SL shows a higher performance than the classic CRO-SL in general. However, it is the DPCRO-SL approach the ensemble which shows the best results over the other CRO-SL versions, with a very significant overall improvement in performance for all test functions. These results indicate that the DPCRO-SL makes a more efficient management of the search resources in the ensemble, by means of modifying the probability of each search procedure as the algorithms evolves.

\begin{table}[!ht]
    \scriptsize
    \centering
    \resizebox{\textwidth}{!}{
    \begin{tabular}{|l|l|l|l|l|l|l|l|l|l|}
    \hline
        Function&\multicolumn{3}{|c|}{DPCRO-SL} &  \multicolumn{3}{|c|}{PCRO-SL} &  \multicolumn{3}{|c|}{CRO-SL}\\ 
 \hline
        \# & best & mean & std & best & mean & std & best & mean & std \\ \hline
        F1 & 4.16E-78 & \textbf{3.20E-76} & 4.99E-76 & 7.60E-62 & 1.20E-60 & 1.95E-60 & 1.91E-68 & 8.07E-58 & 1.91E-57 \\ \hline
        F2 & 2.63E-77 & \textbf{6.86E-75} & 1.29E-74 & 2.98E-61 & 7.20E-60 & 9.00E-60 & 5.07E-63 & 1.19E-56 & 3.57E-56 \\ \hline
        F3 & 1.75E-72 & \textbf{1.42E-69} & 3.12E-69 & 1.83E-55 & 1.05E-54 & 1.45E-54 & 3.22E-61 & 2.26E-48 & 6.03E-48 \\ \hline
        F4 & 1.63E-81 & \textbf{1.42E-78} & 2.25E-78 & 4.00E-64 & 3.43E-63 & 3.41E-63 & 1.68E-68 & 5.00E-54 & 1.50E-53 \\ \hline
        F5 & 2.34E-16 & \textbf{1.49E-10} & 4.44E-10 & 8.98E-13 & 3.99E-01 & 1.20E+00 & 1.19E-15 & 6.63E-08 & 1.33E-07 \\ \hline
        F6 & 3.55E-15 & \textbf{3.55E-15} & 0.00E+00 & 3.55E-15 & \textbf{3.55E-15} & 0.00E+00 & 3.55E-15 & \textbf{3.55E-15} &  0.00E+00 \\ \hline
        F7 & -6.00E+01 & \textbf{-6.00E+01} & 2.66E-14 & -6.00E+01 & \textbf{-6.00E+01} & 3.03E-14 & -6.00E+01 & \textbf{-6.00E+01} & 3.87E-14 \\ \hline
        F8 & 0.00E+00 & 2.71E-03 & 4.33E-03 & 0.00E+00 & \textbf{7.40E-04} & 2.22E-03 & 0.00E+00 & 1.97E-03 & 4.09E-03 \\ \hline
        F9 & 1.39E+01 & 5.73E+01 & 5.57E+01 & 1.79E+01 & \textbf{3.24E+01} & 2.54E+01 & 1.49E+01 & 4.99E+01 & 5.48E+01 \\ \hline
        F10 & 0.00E+00 &\textbf{0.00E+00} & 0.00E+00 & 0.00E+00 & \textbf{0.00E+00} & 0.00E+00 & 0.00E+00 & \textbf{0.00E+00} & 0.00E+00 \\ \hline
        F11 & 0.00E+00 & \textbf{0.00E+00} & 0.00E+00 & 0.00E+00 & \textbf{0.00E+00} & 0.00E+00 & 0.00E+00 & \textbf{0.00E+00} & 0.00E+00 \\ \hline
        F12 & -4.06E+02 & \textbf{-4.06E+02} & 5.68E-14 & -4.06E+02 & \textbf{-4.06E+02} & 5.68E-14 & -4.06E+02 & \textbf{-4.06E+02} & 5.68E-14 \\ \hline
        F13 & 4.94E-01 & \textbf{5.18E-01} & 1.37E-02 & 4.92E-01 & \textbf{5.18E-01} & 1.45E-02 & 4.98E-01 & 5.20E-01 & 1.25E-02 \\ \hline
        F14 & 2.20E+08 & \textbf{2.20E+08} & 1.65E-01 & 2.20E+08 & \textbf{2.20E+08} & 2.18E-01 & 2.20E+08 & \textbf{2.20E+08} & 4.99E-01 \\ \hline
        F15 & 3.47E-01 & \textbf{3.47E-01} & 1.98E-08 & 3.47E-01 & 3.47E-01 & 3.62E-08 & 3.47E-01 & 3.47E-01 & 2.33E-08 \\ \hline
    \end{tabular}
    }
    \caption{Comparison among different CRO-SL ensemble variants (CRO-SL, PCRO-SL and DPCRO-SL) with 4 DE-based substrates.}
    \label{tab:tableCROVAR}
\end{table}

We extend the experiments in the benchmark functions considered by carrying out a comparison with two existing meta-heuristics approaches that have obtained excellent performance in previous works. Specifically, we compare the DPCRO-SL with a version of the PSO algorithm \cite{wang2018particle} and the LSHADE algorithm \cite{wang2021shade}. Table \ref{tab:tableCOMPCRO} shows the results obtained in this comparison. As can be seen in this table, the DPCRO-SL is able to equal or improve the performance of PSO and LSHADE approaches in all cases, and in the most difficult benchmark functions, the differences are significant.

\begin{table}[!ht]
    \scriptsize
    \centering
    \resizebox{\textwidth}{!}{
    \begin{tabular}{|l|l|l|l|l|l|l|l|l|l|}
    \hline
        Function&\multicolumn{3}{|c|}{DPCRO-SL} &  \multicolumn{3}{|c|}{PSO} &  \multicolumn{3}{|c|}{LSHADE}\\ \hline
        \# & best & mean & std & best & mean & std & best & mean & std \\ \hline
        F1 & 4.16E-78 & \textbf{3.20E-76} & 4.99E-76 & 4.98E-33 & 5.98E-31 & 9.51E-31 & 1.21E-79 & 1.90E-73 & 5.70E-73 \\ \hline
        F2 & 2.63E-77 & \textbf{6.86E-75} & 1.29E-74 & 2.48E+04 & 1.72E+05 & 9.93E+04 & 4.14E-73 & 1.60E-67 & 4.56E-67 \\ \hline
        F3 & 1.75E-72 & \textbf{1.42E-69} & 3.12E-69 & 2.24E-25 & 8.00E+03 & 4.00E+03 & 2.05E-72 & 3.89E-66 & 1.04E-65 \\ \hline
        F4 & 1.63E-81 & \textbf{1.42E-78} & 2.25E-78 & 5.27E-33 & 6.29E+01 & 4.86E+01 & 8.53E-77 & 3.61E-73 & 6.33E-73 \\ \hline
        F5 & 2.34E-16 & \textbf{1.49E-10} & 4.44E-10 & 1.25E-01 & 2.02E+05 & 3.99E+05 & 6.13E-03 & 8.84E-01 & 9.16E-01 \\ \hline
        F6 & 3.55E-15 & \textbf{3.55E-15} & 0.00E+00 & 7.11E-15 & 1.21E-14 & 3.26E-15 & 3.55E-15 & 3.91E-15 & 1.07E-15 \\ \hline
        F7 & -6.00E+01 & \textbf{-6.00E+01} & 2.66E-14 & -6.00E+01 & \textbf{-6.00E+01} & 2.59E-14 & -6.00E+01 & \textbf{-6.00E+01} & 0.00E+00 \\ \hline
        F8 & 0.00E+00 & 2.71E-03 & 4.33E-03 & 0.00E+00 & 1.28E-02 & 1.06E-02 & 0.00E+00 & \textbf{0.00E+00} & 0.00E+00 \\ \hline
        F9 & 1.39E+01 & 5.73E+01 & 5.57E+01 & 6.77E+01 & 1.35E+02 & 4.08E+01 & 1.35E-06 & \textbf{4.76E-05} & 1.10E-04 \\ \hline
        F10 & 0.00E+00 & \textbf{0.00E+00} & 0.00E+00 & 0.00E+00 & 2.21E+02 & 2.45E+02 & 0.00E+00 & \textbf{0.00E+00} & 0.00E+00 \\ \hline
        F11 & 0.00E+00 & \textbf{0.00E+00} & 0.00E+00 & 0.00E+00 & \textbf{0.00E+00} & 0.00E+00 & 0.00E+00 & \textbf{0.00E+00} & 0.00E+00 \\ \hline
        F12 & -4.06E+02 & -4.06E+02 & 5.68E-14 & -4.06E+02 & -4.06E+02 & 5.68E-14 & -2.04E+03 & \textbf{-2.04E+03} & 2.27E-13 \\ \hline
        F13 & 4.94E-01 & 5.18E-01 & 1.37E-02 & 4.90E-01 & 4.93E-01 & 2.08E-03 & 4.96E-01 & \textbf{5.03E-01} & 3.58E-03 \\ \hline
        F14 & 2.20E+08 & \textbf{2.20E+08} & 1.65E-01 & 2.20E+08 & 4.56E+08 & 2.88E+08 & 2.20E+08 & \textbf{2.20E+08} & 2.68E-01 \\ \hline
        F15 & 3.47E-01 & \textbf{3.47E-01} & 1.98E-08 & 7.40E-01 & 7.44E-01 & 2.60E-03 & 4.73E-01 & 4.75E-01 & 5.65E-03 \\ \hline
    \end{tabular}
    }
    \caption{Comparison of the DPCRO-SL ensemble with PSO \cite{wang2018particle} and LSHADE algorithm \cite{wang2021shade}.}
    \label{tab:tableCOMPCRO}
\end{table}

\subsection{Comparison in a real problem of wind turbine assignment}

To further test the performance of the proposed DPCRO-SL approach, we have tackled a case study of wind turbines assignment, a challenge described in~{\cite{baker2019best}}. This challenge was proposed by National Renewable Energy Lab (NREL) in the US, together with IEA (International Energy Agency) Wind Task 37, as a case competition in 2019. A circular symmetry wind farm is considered in this problem, on flat and level terrain. The wind turbines $(x, y)$ locations are restricted to be on or within the boundary radius of the wind farm. A separation constraint between turbines is also taken into account (a minimum distance of 2 rotor diameters between turbines is considered).

\begin{figure}[htpb]
    \centering
    \begin{subfigure}[b]{0.48\linewidth}
        \includegraphics[width=1.0\textwidth]{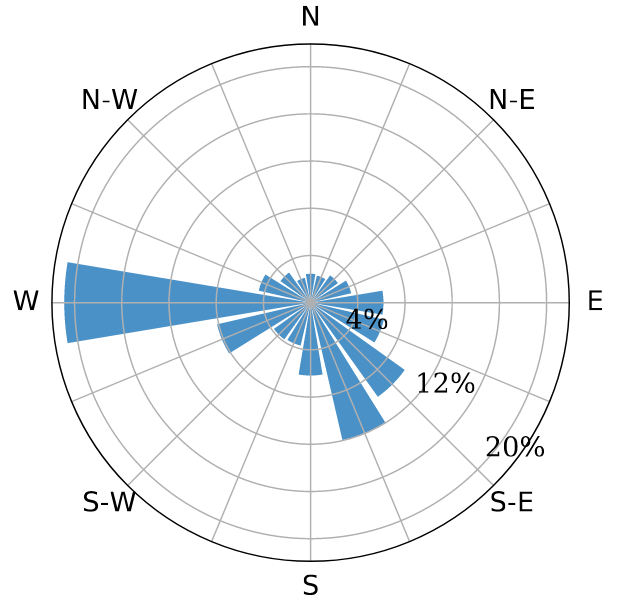}
        \caption{}
        \label{fig:WindRoseChall}
    \end{subfigure}
    \begin{subfigure}[b]{0.48\linewidth}
        \includegraphics[width=1.0\textwidth]{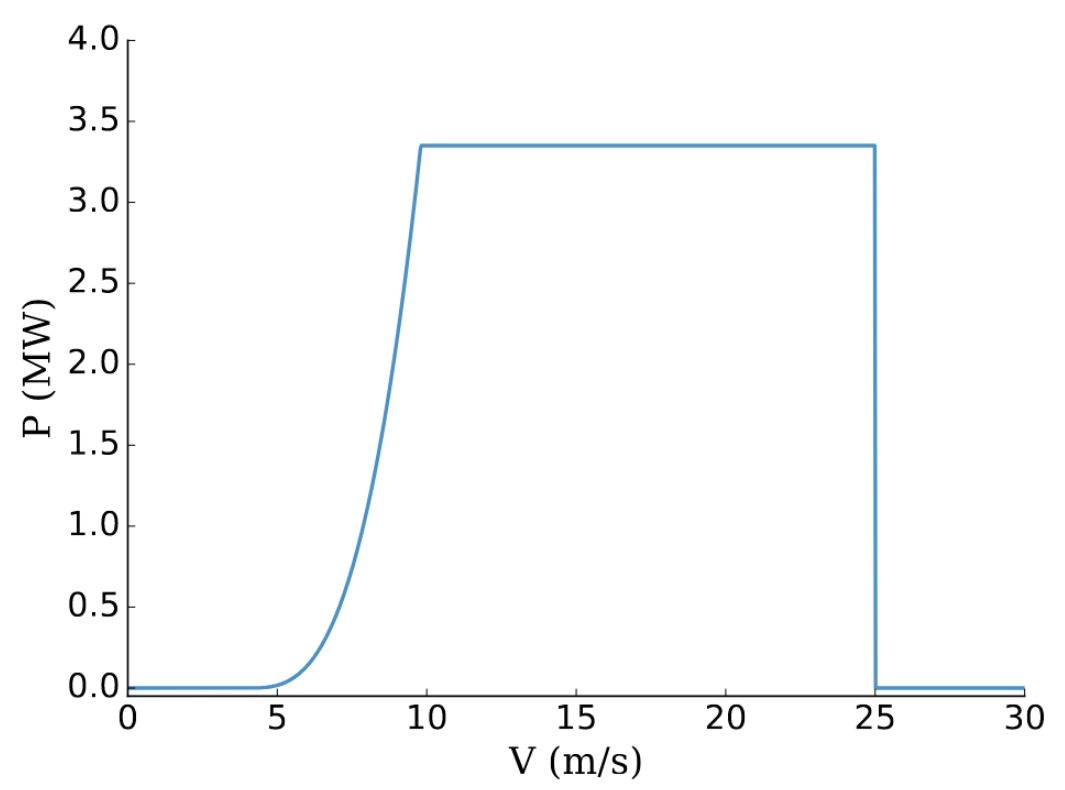}
        \caption{}
        \label{fig:TurbChall}
    \end{subfigure}
    \caption{{\emph{a)} Wind rose of the challenge~{\cite{baker2019best}}. \emph{b)} Power Curve of the NREL's 3.35-MW onshore reference turbine~{\cite{baker2019best}}.}}
    \label{fig:WindRoseTurbChall}
\end{figure}

The wind characteristics considered for this challenge are specified in \cite{baker2019best} and also described in \cite{perez2022versatile}. Briefly, the wind distribution frequency and wind speed are the same for all wind farm scenarios. Free-stream wind velocity is constant in all wind directions, fixed to $9.8$ m/s, for all days. The challenge considers a wind rose (Figure ~\ref{fig:WindRoseChall})  with an off-axis wind frequency distribution, binned for $16$ directions. 

Regarding the turbines characteristics, the case study considers the use of the IEA's 3.35-MW reference turbine. Its attributes are open source, and it is designed as a baseline for onshore wind turbine specifications~{\cite{bortolotti2019iea}}. The specifics of the turbine are shown in Table~\ref{tab:chaTurb}.

\begin{table}[htpb]
    \centering
    \begin{tabular}{l|c|c}
    \hline\hline
        \textbf{Parameter} & \textbf{Value} & \textbf{Units}\\\hline
        Rotor Diameter & 130&  m\\
        Turbine Rating&  3.35&  MW\\
        Cut-In Wind Speed&  4&  m/s\\
        Rated Wind Speed&  9.8&  m/s\\
        Cut-Out Wind Speed&  25&  m/s\\\hline\hline
    \end{tabular}
    \caption{{Attributes for NREL's 3.35-MW onshore reference turbine~{\cite{bortolotti2019iea}}.}}
    \label{tab:chaTurb}
\end{table}

{Figure~{\ref{fig:TurbChall}} shows the turbine power curve considered.}
{and Equation~{\eqref{eq:turbChall}} provides its analytic expression:}
\begin{equation}\label{eq:turbChall}
    P(V) = \left \lbrace
    \begin{aligned}
    &0 & V<V_{cut-in}\\
    &P_{rated} \left( \frac{V -V_{cut-in}}{V_{rated} -V_{cut-in} }\right)^3 & V_{cut-in} < V < V_{rated}\\
    &P_{rated} & V_{rated} < V < V_{cut-out} \\
    &0 & V_{cut-out} < V
    \end{aligned}\right .
\end{equation}

% % wind turbine round layout

% The challenge specifies three different sub-cases or scenarios that basically differ from the size of the circular domain and the number of turbines involved:
% \begin{enumerate}
%     \item {Scenario 1: a wind farm boundary radius of 1300 m with 16 turbines.}
%     \item {Scenario 2: a wind farm boundary radius of 2000 m with 36 turbines.}
%     \item {Scenario 3: a wind farm boundary radius of 3000 m with 64 turbines.}
% \end{enumerate}

We consider here the first scenario of the challenge, consisting in a wind farm boundary radius of 1300m with 16 turbines to be positioned. The metric used in this challenge is the annual energy production (AEP) for the turbine layout which has the following expression:
\begin{equation}
  \mathit{AEP} = \left(\sum_{i=1}^m f_iP_i\right)8760\text{ } \frac{\text{hrs}}{\text{yr}},
\end{equation}
where $f_i$ is the corresponding frequency for the direction $i$ and $P_i$ is the wind farm power for direction $i$. Note that $8760$ are the number of hours in a year.

\subsubsection{Results}

In this case we evaluate the performance of the DPCRO-SL, considering five substrates in the search, DE/best/1, Fa, BLX, GM and CM. A local search given by a Cauchy-based mutation has also been applied. 

Table \ref{tab:scenario1Chall} shows the results obtained with the DPCRO-SL, and a comparison of performance between the DPCRO-SL and alternative approaches in the literature (from \cite{baker2019best}). We have classified the different approaches in Gradient-based (G) or Gradient Free (GF) algorithms. As can be seen, the best performance in this problem has been obtained by the proposed DPCRO-SL, with a best AEP of 419935.8, following from different gradient-based approaches (see \cite{baker2019best} for details on these approaches). Note that alternative meta-heuristics such as PSO or evolutionary algorithms are far away from the DPCRO-SL performance in this problem. 

\begin{table}[!ht]
    \centering
    \begin{tabular}{rlcr}
    \hline\hline
         Rank & Algorithm & Grad. & AEP \\\hline
         \textbf{1} & \textbf{DPCRO-SL} & \textbf{GF} & $\mathbf{419935.7905}$\\
         2 &  SNOPT+WEC &  G & $418924.4064$ \\
         3 & fmincon &  G &  $414141.2938$\\
         4 &  SNOPT &  G &  $412251.1945$ \\
         5 &  SNOPT &  G &  $411182.2200$ \\
         6 &  PSQP &  G &  $409689.4417$ \\
         7 &  Multistart Interior-Point &  G & $408360.7813$ \\
         8 &  Full Pseudo-Gradient Approach & GF &  $402318.7567$\\
         9 &  Basic Genetic Algorithm  & GF & $392587.8580$  \\
         10 &  Simple Particle Swarm Optimization & GF & $388758.3573$ \\
         11 & Simple Pseudo-Gradient Approach & GF & $388342.7004$ \\
         \\\hline\hline
    \end{tabular}
    \caption{{Results in a wind farm boundary radius of 1300 m with 16 turbines.}}
    \label{tab:scenario1Chall}
\end{table}

\begin{figure}[!ht]
    \centering
    \includegraphics[width=0.9\linewidth]{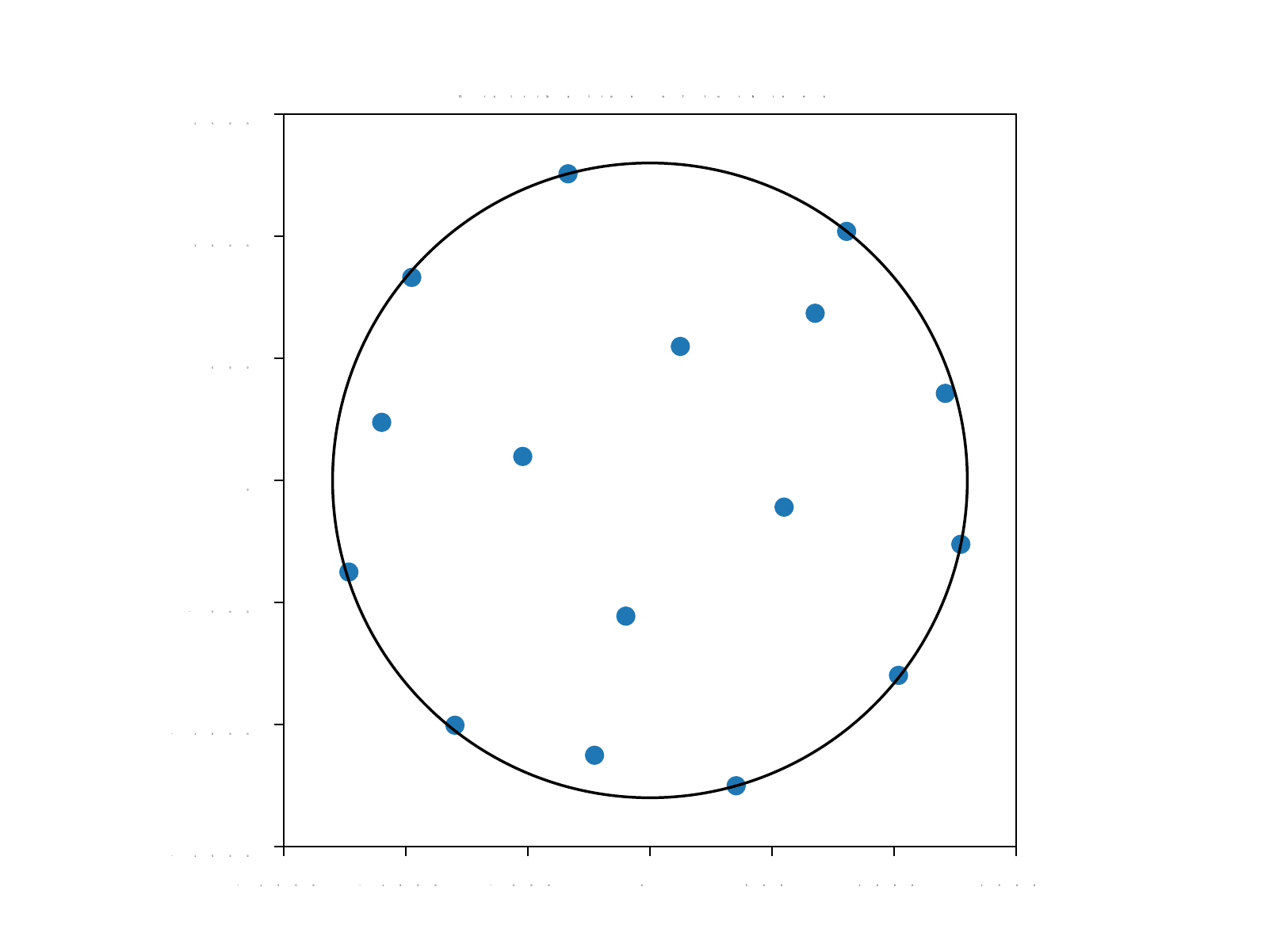}
    \caption{
    {Turbines layout by means of the proposed DPCRO-SL (16 turbines, 1300m radius).
    }}
    \label{fig:LayoutChall1}
\end{figure}

Figure \ref{fig:LayoutChall1} shows the best layout obtained in the problem, with the DPCRO-SL. This solution is also shown in Table \ref{tab:CRO-SLScenario1}. As can be seen, this best solution spreads all the possible wind turbines at the edge of the wind farm, with almost a regular separation. The rest of turbines are distributed over the center of the wind farm, with enough distance among them.\\ 

% \begin{figure}[!ht]
%     \centering
%     \includegraphics[width=0.9\linewidth]{turb_plots/fit_turb16_17.pdf}
%     \caption{
%     {Turbines layout by means of the proposed DPCRO-SL (16 turbines, 1300m radius).
%     }}
%     \label{fig:LayoutChall1}
% \end{figure}

\begin{table}[htpb]
    \centering
    \begin{tabular}{c|c|c|c|c|c|c|c|c}
    \hline\hline
        i & 1 & 2 & 3 & 4 & 5 & 6 & 7& 8\\\hline
         x & -335.6 & 1273.3 & 1210.0 & -521.1 & -798.7 & -226.9 & 124.6 & 1018.1  \\
         y & 1255.7 & -261.8 & 356.3 & 98.0 & -1003.0 & -1125.9 & 548.6 & -798.7\\\hline
         i & 9 & 10 & 11 & 12 & 13 & 14 & 15& 16\\\hline
         x & -1233.3 & -975.6 & 805.6 & 676.7 & -1098.8 & 549.4 & 353.1 & -98.7\\
         y & -375.5 & 831.4 & 1019.8 & 684.4 & 237.8 & -109.7 & -1250.9 & -556.0\\\hline\hline
    \end{tabular}
    \caption{{Best turbine layout solution obtained by the DPCRO-SL (16 turbines, 1300m radius).}}
    \label{tab:CRO-SLScenario1}
\end{table}

\newpage
\section{Conclusions}\label{sec:Conclusions}
In this paper we have proposed two new probabilistic and dynamic multi-method ensembles, based on the Coral Reefs Optimization with Substrate Layer (CRO-SL). Specifically, we have first defined a probabilistic CRO-SL, where the classical substrates of the original algorithm are changed by tags associated with each coral (solution), which points out to a given search method. In this version, the tags which relate solutions and search methods are changed in every generation of the algorithm, leading to a probabilistic version of the ensemble, in contrast to the original static CRO-SL version. The second proposed multi-method ensemble is a dynamic version, where the probability of the tags assignment varies during the evolution of the algorithm, depending on the performance of the search methods in the problem at hand. We have tested the performance of the proposed multi-method ensembles in different optimization problems, including different benchmark functions and a real problem of wind turbines layout. Comparison with state of the art algorithms has shown an excellent performance of the proposed ensembles, mainly for the dynamic probabilistic version of the CRO-SL. These good results show that the novel multi-method ensembles proposed are potentially excellent algorithms for a high number of optimization problems, including real-world optimization tasks. Finally, associated with this paper, we provide free access to the Python code of the DPCRO-SL, via github, so any researcher can download the code, modify it, add new search strategies, and test it in any other optimization problem.

\section*{Acknowledgments}
This research has been partially supported by the project PID2020-115454GB-C21 of the Spanish Ministry of Science and Innovation (MICINN)

\section*{Code availability} 
DPCRO-SL code in Python available at: \url{https://github.com/jperezaracil/PyCROSL.git}
%Bibliography

\section*{Appendices}

\subsection*{Benchmark functions}\label{secA1}
\begin{itemize}
    \item F1: Sphere
    
    \begin{equation*}
        f_1(\bm{x}) = \sum^N_{i=1} {x}^2_i
    \end{equation*}
    
    \item F2: High Condition Elliptic

    \begin{equation*}
        f_2(\bm{x}) = \sum^N_{i=1} 10^{6\frac{i-1}{N-1}}{x}^2_i
    \end{equation*}
    
    \item F3: Bent Cigar
    \begin{equation*}
        f_3(\bm{x}) = {x}_1^2 + 10^6\sum^N_{i=2} {x}^2_i
    \end{equation*}
    
    \item F4: Discus
    \begin{equation*}
      f_4(\bm{x}) = 10^6{x}_1^2 + \sum^N_{i=2} {x}^2_i  
    \end{equation*}
    
    \item F5: Rosenbrock
    \begin{equation*}
        f_5(\bm{x}) = \sum^{N-1}_{i=1} 100({x}_{i+1}-{x}_i^2)^2 + (1 - {x}_1)^2
    \end{equation*}
    
    \item F6: Ackley
    \begin{equation*}
        f_6(\bm{x}) = e - 20exp{\left(-0.2 \cdot \sqrt{\sum^N_{i=1} {x}_i^2}\right)} -
        exp{\left(\frac{1}{N} \sum^N_{i=1} cos(2 \pi {x}_i)\right)} + 20
    \end{equation*}
    
    \item F7: Weierstrass (limited to 20 iterations)
    \begin{equation*}
        f_7(\bm{x}) = \sum^{N}_{i=1} \sum^{20}_{j=1} 0.5^{j} \cdot cos(2\pi \cdot 3^{j} \cdot ({x}_i + 0.5))
    \end{equation*}
    
    \item F8: Griewank
    \begin{equation*}
        f_8(\bm{x}) = 1 + \frac{1}{4000}\sum^N_{i=1} {x}^2_i - \prod^{n}_ {i=1}cos \left(\frac{{x}_i}{\sqrt{i}}\right)
    \end{equation*}
    
    \item F9: Rastrigin
    \begin{equation*}
        f_9(\bm{x}) = 10N + \sum^N_{i=1} \left[{x}^2_i - 10 cos(2\pi {x}_i)\right]
    \end{equation*}
    
    \item F10: Modified Schwefel
    \begin{equation*}
        \begin{aligned}
            g_{10}(x) &= 
            \begin{cases}
                - x \, sin\left(\sqrt{x}\right) & x = 500\\
                - \left(500 - [x \, mod \, 500] \cdot sin\left(\sqrt{500 - [x \, mod \, 500]}\right)\right) + \left(\frac{x-500}{N^2 100}\right)^2 & x > 500\\
                - \left(-500 - [x \, mod \, 500] \cdot sin\left(\sqrt{500 - [x \, mod \, 500]}\right)\right) + \left(\frac{x+500}{N^2 100}\right)^2 & x < 500\\
            \end{cases}\\            
            f_{10}(\bm{x}) &= N \sum^{N}_{i=1} g_{10}({x}_i) 
        \end{aligned}
    \end{equation*}
    
    \item F11: Katsuura
    \begin{equation*}
        f_{11}(\bm{x}) = \frac{10}{N^2}\prod^N_{i=1} \left[1 + (i+1)\sum^{N}_{k=1}\lfloor 2^k x_i\rfloor 2^{-k} \right]
    \end{equation*}
    
    \item F12: Happy Cat
    \begin{equation*}
        f_{12}(\bm{x}) = (\|\bm{x}\|^2 - N)^{0.25} + \frac{1}{N} \left(\frac{1}{2}\|\bm{x}\|^2 + \sum^N_{i=1} {x}_i\right) + \frac{1}{2}
    \end{equation*}
    
    \item F13: HGBat
    \begin{equation*}
        f_{13}(\bm{x}) = \left(\|\bm{x}\|^4 - \left(\sum^N_{i=1} {x}_i\right)^{2}\right)^{0.25} + \frac{1}{N} \left(\frac{1}{2}\|\bm{x}\|^2 + \sum^N_{i=1} {x}_i\right) + \frac{1}{2}
    \end{equation*}

    \item F14: Griewank plus Rosenbrock
    \begin{equation*}
    \begin{aligned}
        g_{14}(\bm{x}) &=
        \begin{split}
         \sum^{N-1}_{i=1} & \left[\frac{1}{4000}(100({x}_{i}^2-{x}_{i+1}) + ({x}_{i}-1)^2)^2 - \right.\\ & \left. cos(100({x}_{i}^2-{x}_{i+1}) + ({x}_{i}-1)^2) + 1 \right]
        \end{split}\\
        f_{14}(\bm{x}) &=
              g_{14}(\bm{x}) + \frac{1}{4000}f_{5}(\bm{x})^2 - cos(f_{5}(\bm{x})) + 1
    \end{aligned}
    \end{equation*}
    
% \begin{equation}
% \begin{alignedat}{2}
%     \label{eq:IsoLaplaceFinal}
%     \boldsymbol{Y}_I(s)=\left[\boldsymbol{I}_n-\boldsymbol{G}_{IB}(s)\boldsymbol{G}_{BI}(s) \right]^{-1}\boldsymbol{G}_{IB}(s)\boldsymbol{G}_{BD}(s)\boldsymbol{F}_d(s)+ \\ 
%     \left[\boldsymbol{I}_n-\boldsymbol{G}_{IB}(s)\boldsymbol{G}_{BI}(s) \right]^{-1}\boldsymbol{G}_{IF}(s)\boldsymbol{F}_a(s),
%     \end{alignedat}
% \end{equation}
% 

% \begin{alignat}{2}
% A_{0} & = -\frac{3\sigma_{1}' h H L^{2}}{E (H+h)^{3}} &{}={}& -1.62\times 10^{-6}\,\mathrm{m} \\
% B_{0} & = - \frac{3 h H (\alpha_{2}-\alpha_{1}) L^{2}}{(H+h)^{3}} &{}={}&  2.95\times 10^{-8}\,\mathrm{m\,K^{-1}} \\
% C_{0} & = \frac{4 L^{3}}{E W (H+h)^{3}} &{}={}& 1.08\,\mathrm{m\,N^{-1}}
% \end{alignat}
    
    \item F15: Exp Shaffer F6
    \begin{equation*}
        f_{15}(\bm{x}) = 1 + \frac{sin^2\left(\sqrt{\sum^{N-1}_{i=1} ({x}^2_i + {x}^2_{i+1})}\,\right) - 0.5}{\left[1+0.001\sum^{N-1}_{i=1} ({x}^2_i + {x}^2_{i+1})\right]^2}
        + \frac{sin^2\left(\sqrt{(N-1)^2 + {x}_1^2}\,\right) - 0.5}{\left[1+0.001((N-1)^2 + {x}_1^2)\right]^2}
    \end{equation*}
    
\end{itemize}

\bibliographystyle{unsrt}  
\bibliography{templateArxiv}

\end{document}